\documentclass[letterpaper, 10 pt, conference]{ieeeconf}

\IEEEoverridecommandlockouts

\usepackage{graphics} %
\usepackage{graphicx}
\usepackage{epsfig} %
\usepackage{amsmath} %
\usepackage{amssymb}  %
\usepackage{subfigure}
\usepackage{float}
\usepackage{url}
\usepackage{multirow}
\usepackage{subfigure}
\usepackage{wrapfig}
\usepackage{tabularx}
\usepackage{hyperref} %
\usepackage{color}
\usepackage{bm}
\usepackage{array}
\usepackage{booktabs}
\usepackage{algorithm,algorithmic}
\usepackage{siunitx}

\usepackage{gensymb}

\usepackage{lipsum}
\usepackage{makecell}

\graphicspath{{./figs/}}

\newcolumntype{C}[1]{>{\centering}m{#1}}
\newcolumntype{L}{>{\centering\arraybackslash}m{0.7cm}}
\makeatletter
\newcommand{\thickhline}{%
    \noalign {\ifnum 0=`}\fi \hrule height 1pt
    \futurelet \reserved@a \@xhline
}
\newcolumntype{"}{@{\hskip\tabcolsep\vrule width 1pt\hskip\tabcolsep}}
\makeatother

\newcommand{\doctitle}{I Can See Clearly Now : Image Restoration via De-Raining}
\newcommand{\docsubtitle}{}
\title{\LARGE \bf\doctitle \small\break\docsubtitle}
\author{Horia Porav, Tom Bruls and Paul Newman
\thanks{Authors are from the Oxford Robotics Institute, University of Oxford, UK.
{\tt\small \{horia,tombruls,pnewman\}@robots.ox.ac.uk}}
}

\renewcommand{\baselinestretch}{0.95}

\begin{document}
\maketitle
\thispagestyle{empty}
\pagestyle{empty}

\begin{abstract}

We present a method for improving segmentation tasks on images affected by adherent rain drops and streaks. We introduce a novel stereo dataset recorded using a system that allows one lens to be affected by real water droplets while keeping the other lens clear. We train a denoising generator using this dataset and show that it is effective at removing the effect of real water droplets, in the context of image reconstruction and road marking segmentation. To further test our de-noising approach, we describe a method of adding computer-generated adherent water droplets and streaks to any images, and use this technique as a proxy to demonstrate the effectiveness of our model in the context of general semantic segmentation. We benchmark our results using the CamVid road marking segmentation dataset, Cityscapes semantic segmentation datasets and our own real-rain dataset, and  show significant improvement on all tasks.
\end{abstract}

\section{Introduction}\label{sec:introduction}

If we want machines to work outdoors and see while doing so, they have to work in the rain. When rain and lenses interact, computer vision becomes harder - wild local distortions of the image appear which dramatically impede image understanding tasks. However the distortions are not noise, they are structured, the light field is simply bent and attenuated,  and accordingly can be modelled and reversed. 

In this work we develop a filter which as a pre-processing step removes the effect of raindrops on lenses. Several tasks are affected by the presence of adherent water droplets on camera lenses or enclosures, such as semantic segmentation \cite{cordts2016cityscapes}, localisation using  segmentation \cite{stenborg2018,sattler2018semloc} or road marking segmentation \cite{bruls2018mark}. In this paper we choose to use  segmentation as an example task by which to test the effectiveness of our method. Many approaches so far have reached for multi-modal data \cite{adapnet}, domain adaptation \cite{vangool2018,Wulfmeier2017} or  training on synthetic data \cite{Synthia}, however this can become awkward as:
\begin{enumerate}
    \item Acquiring rainy images is time-consuming, expensive or impossible for many tasks or setups, especially in the case of supervised training, where ground truth data is needed.
    \item Training, domain-adapting or fine-tuning each individual task with augmented data is intractable.
\end{enumerate}

\begin{figure}[!htbp]
\centering

{\includegraphics[width=0.498\columnwidth]{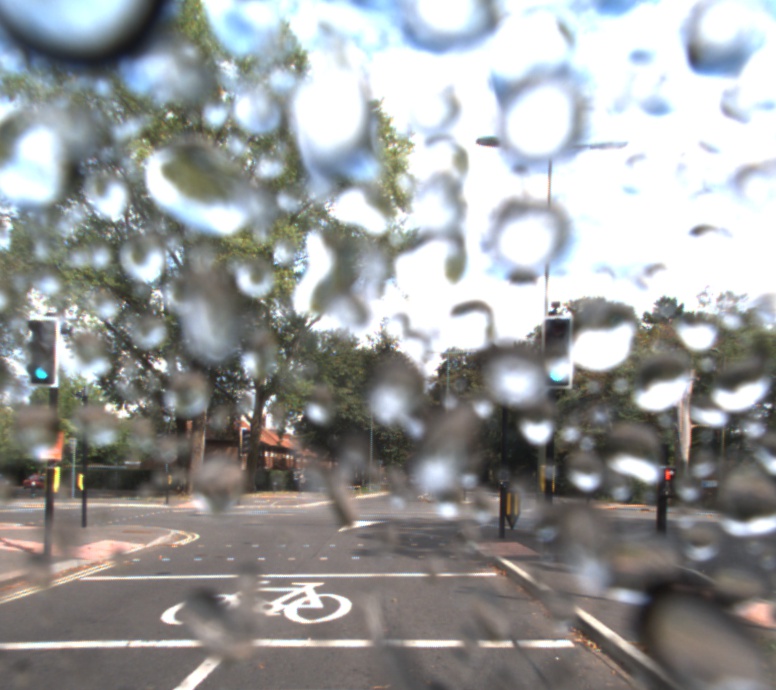}}%
\hfill %
{\includegraphics[width=0.498\columnwidth]{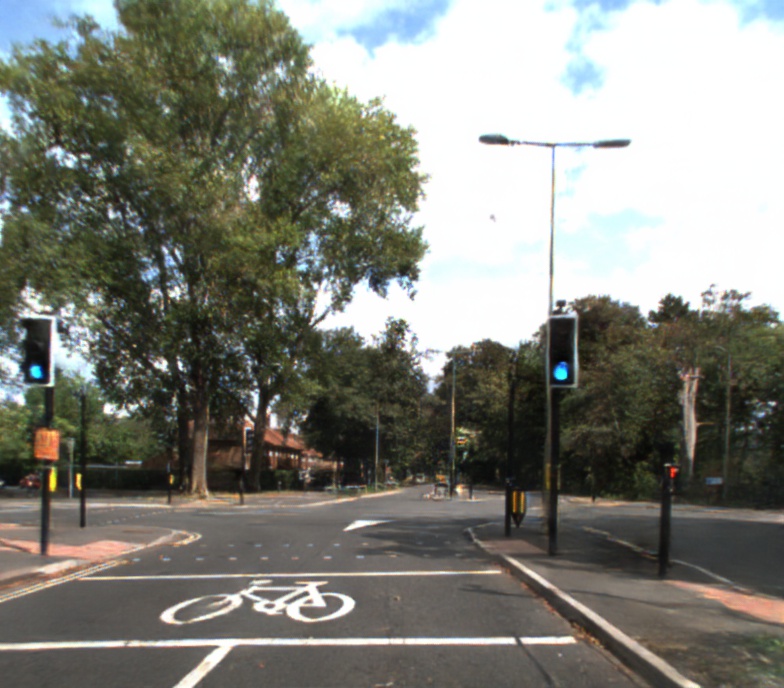}}%
\\ %
\vspace{0.5mm}

{\includegraphics[width=0.498\columnwidth]{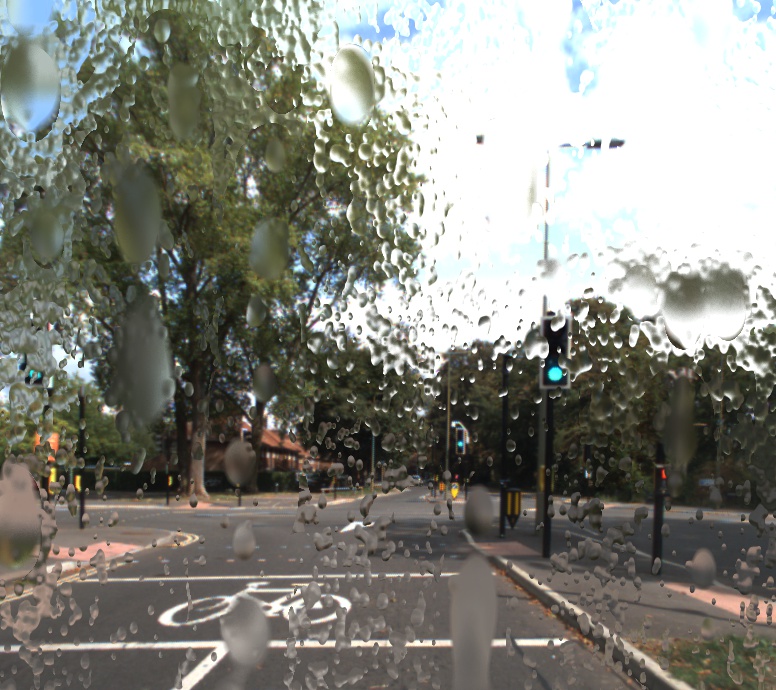}}%
\hfill %
{\includegraphics[width=0.498\columnwidth]{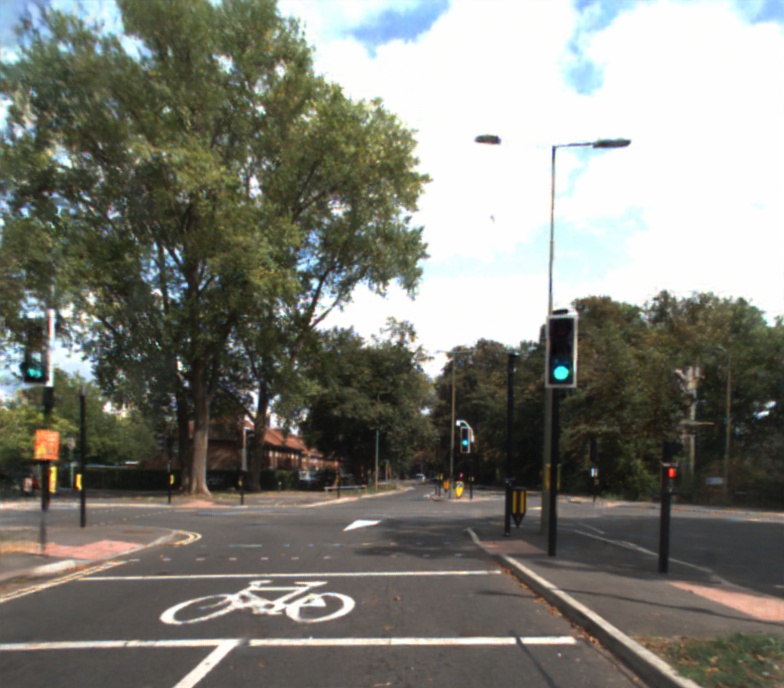}}%

\caption{We learn a de-noising generator that can remove noise and artefacts induced by the presence of adherent rain droplets and streaks. On the top left,input images that are affected by real rain drops. On the top right, the cleaned, de-rained images. On the bottom left, input images that are affected by computer-generated rain drops. On the bottom right, the cleaned, de-rained images.}
\label{fig:introfig}
\end{figure}

We take a different approach and build a system as an image preprocessor, the output of which is a cleaned, de-rained image that improves the performance of many tasks performed on the image. 

We begin by creating a bespoke real-world small baseline stereo dataset where one lens is affected by real water droplets and the other is kept dry. The methodology and apparatus for doing so is presented in section \ref{sec:rainmaker}. Using this dataset, we train a de-raining generator and show that it is able to both drastically improve the visual quality of images and restore performance on road marking segmentation tasks.

Secondly, we describe a way of efficiently adding computer-generated adherent rain droplets and adherent streaks to any image using GPU shaders. This system is presented in section \ref{subsec:synthrain}. As the Cityscapes dataset provides a good groundtruth for segmentation but does not contain images with significant rain on the lens, we modify it using this technique and use it as a proxy to study the effects of rain on general semantic segmentation. Additionally, we create a synthetic rain dataset by adding computer-generated rain drops to a full Oxford RobotCar dataset \cite{maddern20171} and to the CamVid \cite{brostow2009semantic} dataset.

Our main contributions include: 
\begin{itemize}
    \item a de-raining model that produces state of the art results;
    \item using computer-generated water drops as a proxy to study the effects of rain on segmentation for datasets that provide a ground truth but do not normally contain rainy images; and
    \item a real-world very-narrow-baseline stereo dataset with rainy \& clear images covering a wide array of dynamic scenes.
\end{itemize} 

Our aim is to show that pre-processing the image leads to better performance as compared to training, retraining or fine-tuning a task-specific model with rain-augmented data. We benchmark our de-raining model on the following tasks:

\begin{itemize}
    \item Road marking segmentation and image restoration on a real-world small baseline stereo dataset where one lens is affected by real water droplets and the other is kept dry and clear.
    \item Image reconstruction on the real-world dataset of \cite{Qian2018}.
    \item Road marking segmentation and image restoration on CamVid \cite{brostow2009semantic} and RobotCar \cite{maddern20171} imagery with computer-generated droplets added.
    \item Semantic segmentation on Cityscapes \cite{cordts2016cityscapes} imagery with computer-generated droplets added.
\end{itemize}

The quantitative and qualitative results are presented in section \ref{sec:results}.

\begin{figure}[tp]
\centering
 \includegraphics[trim={0cm 0.5cm 0cm 0cm},width=\columnwidth]{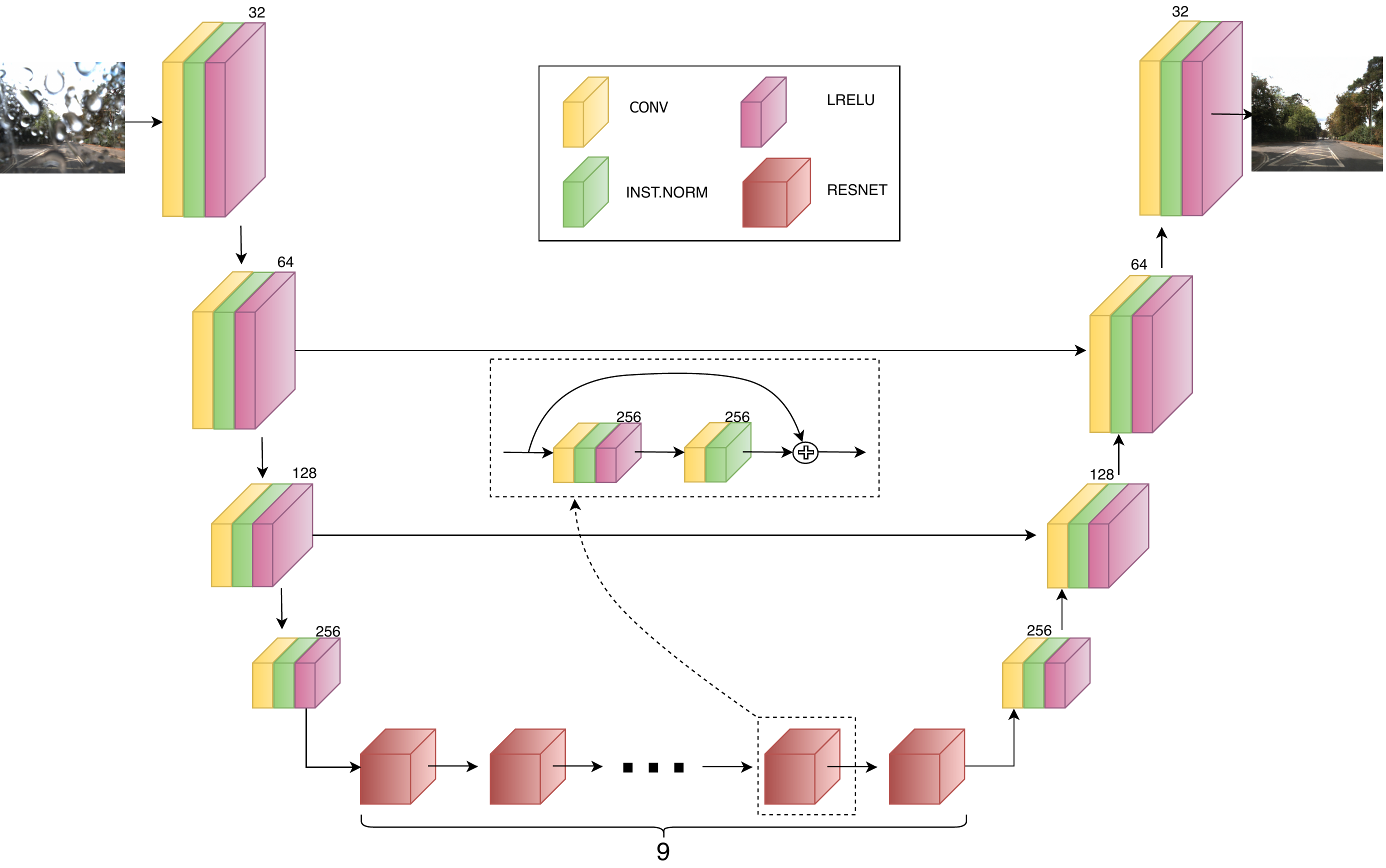}
\caption{The internal architecture of our generator. We motivate the addition of additive skip connections by observing that much of the structure of the input image should be kept, along with illumination levels and fine details.} 
\label{fig:internal}
\end{figure}

\section{Related Work}\label{sec:related-work}

Generally speaking, the quality of an image can be affected in two ways by bad weather conditions. Firstly, contaminants in the atmosphere, such as falling rain, fog, smog or snow will hinder visibility or partially occlude a scene but do not significantly distort the image. Secondly, adherent contaminants such as water droplets, which stick to transparent surfaces or lenses, tend to heavily distort the image, essentially acting as a secondary lens with various degrees of blurring.  Several techniques are employed to clean the first type of images, such as those used by \cite{Chen2014,Kim2014,Kim2015,ren2016single,Fu2016}, however these techniques cannot be used to restore images affected by adherent rain, as the optics involved differ significantly from those of atmospheric droplets. The remainder of this section outlines some of the techniques used to tackle the effects of adherent rain droplets and adherent streaks.

\subsubsection*{\textbf{Rain Modelling and Simulation}}
In the context of computer vision, several studies have attempted to model the structure and optical properties of adherent water droplets. The authors of RIGSEC \cite{GeigerReg2009, geigerModel2011} model raindrops first as sections of a sphere and later account for the effect of gravity using 2D Bezier curves, and confirm experimentally that a physically correct droplet shape can be computed using this method. \cite{waterdropStereo2016} additionally study and model the dark band around the edges of adherent drops, and show that a simplified model is enough to correctly undistort the image on the surface of the droplet.

We base our simple synthetic droplet model on the works of \cite{GeigerReg2009,geigerModel2011} and \cite{waterdropStereo2016}, by storing proto-droplet normal maps which are subsequently warped and combined at run time using an approach similar to meta-balls \cite{Blinn1982}.

Additionally, several small datasets have been created to benchmark the accuracy of de-raining techniques. In \cite{EigenDerain}, water is sprayed on a glass pane fitted in front of a camera, but no ground truth is provided due to temporal illumination and scene changes. A video sequence where the lens is affected by real rain droplets is also provided, again without ground truth.
The authors of \cite{You2016} again use a glass pane sprayed with water to study the performance of their droplet detection and removal pipeline, but only offer ground truth for the position of the droplets. 
The first attempt to provide accurate ground truth is made by \cite{Qian2018}, in which images of static scenes are captured both with and without a glass pane sprayed with water in front of the camera. This process is, however, very difficult to scale to the number of images required by modern deep-learning approaches.
To our best knowledge, we are the first to record a real-world large dataset of sequential dynamic scenes with an accurate, clear ground truth and a large variation in raindrop type and size.
 
\subsubsection*{\textbf{Raindrop Detection and Removal}}
In \cite{GeigerReg2009} and \cite{geigerModel2011}, raindrops are detected by attempting to match a template of a synthetic raindrop at locations where the presence of a real drop is hypothesized. This approach breaks down when the shape of the real droplets differs significantly from that of the template. The authors of \cite{You2016} take a different approach by observing that the motion inside droplets is between 1/30 and 1/20 slower than that in the scene. They use this information to detect raindrops and then attempt to restore the image by using a combination between image inpainting and recovering data from within the distorted image formed on the droplet. Both techniques use multi-frame information for image reconstruction, and are not applicable to single-images. 
 
Multi-camera and pan-tilt setups are exploited by \cite{Yamashita2003,Yamashita2004,Yamashita2005} and \cite{kuramoto2002}. These techniques use disparities to detect droplets and subsequently attempt to replace the affected regions in one lens with information from the other lens. This approach does not work on single images and assumes that the same regions are not covered by rain in both frames.
 
Convolutional neural networks were used by \cite{EigenDerain} to restore images affected by dirt and rain. They use a simple 3-layer architecture, each with 512 units, which works well on small drops but breaks down with much larger contaminants. A much larger Generative Adversarial Network (GAN) model \cite{GAN} is used by \cite{Qian2018}, along with attention \cite{attention}. They leverage their static dataset to provide a ground truth for the droplet attention mask and train a recurrent model that outputs a heatmap of the location of the droplets. This heatmap is then concatenated with the input image and run through the GAN. They produce state-of-the-art results and made their dataset publicly available, which has allowed us to directly compare our method with theirs.

\section{Learning to Clean Images}\label{sec:method}

\subsection{Computer-Generated Synthetic Rain}
\label{subsec:synthrain}

We base our simple synthetic droplet model on the works of \cite{GeigerReg2009,geigerModel2011} and \cite{waterdropStereo2016}, generate the locations of raindrops using a simple statistical approach, model the interactions between raindrops using metaballs \cite{Blinn1982} and implement its rendering efficiently using GPU shaders.

\begin{figure}[h]
 \includegraphics[width=\columnwidth,trim={0cm 0cm 0cm 0cm}]{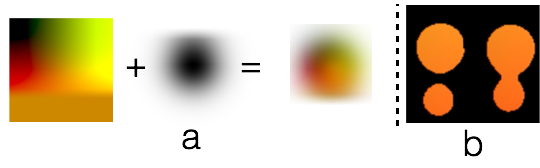}
\caption{Metaballs.}
\label{fig:metaballs}
\end{figure}

A proto-raindrop is created using a simple refractive model that assumes a pinhole camera. The refraction angle is encoded following a scheme similar to normal mapping \cite{normalmap} by using a 2D look-up table represented by the RED and GREEN channels of a texture $T$, with the thickness of the drop encoded in the BLUE channel of the same texture. This texture $T$ is then masked using an alpha layer that allows blending of the water drops with the background image and other drops, as shown in Figure \ref{fig:metaballs}a. With the drop acting as a simple lens, the coordinate $(x_r,y_r)$ of the world point that is rendered at the location $(u,v)$ on the surface of a drop is given by the following simplified distortion model:
\vspace{-1mm}
\begin{equation}
    x_r = u + (R*B)
    \vspace{-2mm}
\end{equation}
\begin{equation}
    y_r = v + (G*B).
\end{equation}
\vspace{-1mm}
Each image location $(u,v)$ has a probability $P_r$ of becoming the center of a proto-raindrop whose dimensions are scaled along the horizontal and vertical directions by a tuple of random values $S_x$ and $S_y$. For each timestep, the center of a droplet may undergo a slip of $D_x$ pixels along the horizontal and $D_y$ pixels along the vertical direction as a function of the droplet diameter $d$:
\vspace{-2mm}
$$
D_x,D_y = \left\{
        \begin{array}{ll}
            0, 0 & \quad d \leq 4mm \\
            x\sim\mathcal N(0,3),P_d * 5 & \quad d > 4mm,
        \end{array}
    \right.
$$
where $P_d$ represents the probability of slip along the vertical direction and $x$ denotes the random deviation of the slip along the horizontal direction.

For each timestep, droplets that are close to each other are merged using the metaballs approach \cite{Blinn1982}, as shown in Figure \ref{fig:metaballs}b. By default, each texture location $T(u,v)$  that does not fall under a droplet encodes a normal that is perpendicular to the background image. Finally, the image is sampled using the normal map defined by the texture $T$ to produce a result similar to the one in the top-left corner of Fig \ref{fig:introfig}.

Using this technique we have created three synthetic rain datasets:
\begin{itemize}
    \item synthetic rain added to CamVid, complete with road marking ground truth;
    \item synthetic rain added to Cityscapes, complete with semantic segmentation ground truth; and
    \item synthetic rain added to the dry images from our stereo dataset, complete with road marking ground truth.
\end{itemize}

\subsection{The de-raining network}\label{subsec:cleaner}

The de-raining network architecture is based on Pix2PixHD \cite{wang2018highresolution}. The architecture is shown in Fig. \ref{fig:internal}. We employ 4 down-convolutional layers with stride 2, followed by 9 ResNet \cite{he2016deep} blocks and 4 up-convolutional layers. We motivate the addition of skip connections by observing that most of the structure of the input image should be kept, along with illumination levels and fine details.

To promote better generalization and inpainting, we refrain from using any direct pixel-wise loss and instead use a combination of adversarial, perceptual, and multi-scale discriminator feature losses. The discriminator architecture is a CNN with 5 layers, similar to PatchGAN \cite{li2016precomputed}.
We present the full structure of the losses in the next section.

\begin{figure*}[!htbp]
\centering
{\includegraphics[width=0.249\textwidth]{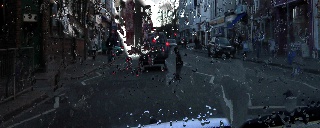}}%
\hfill %
{\includegraphics[width=0.249\textwidth]{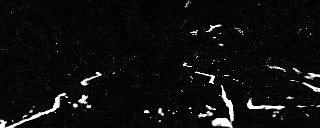}}%
\hfill %
{\includegraphics[width=0.249\textwidth]{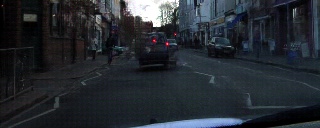}}%
\hfill %
{\includegraphics[width=0.249\textwidth]{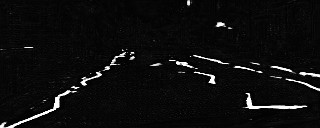}}%
\vspace{-3mm}
\caption{CamVid road marking segmentation results. From left to right: rainy input image, segmentation result on rainy image, derained input image, segmentation result on derained image.}
\label{fig:camvid}
\vspace{-3mm}
\end{figure*}

\begin{figure*}[!htbp]
\centering

{\includegraphics[width=0.248\textwidth]{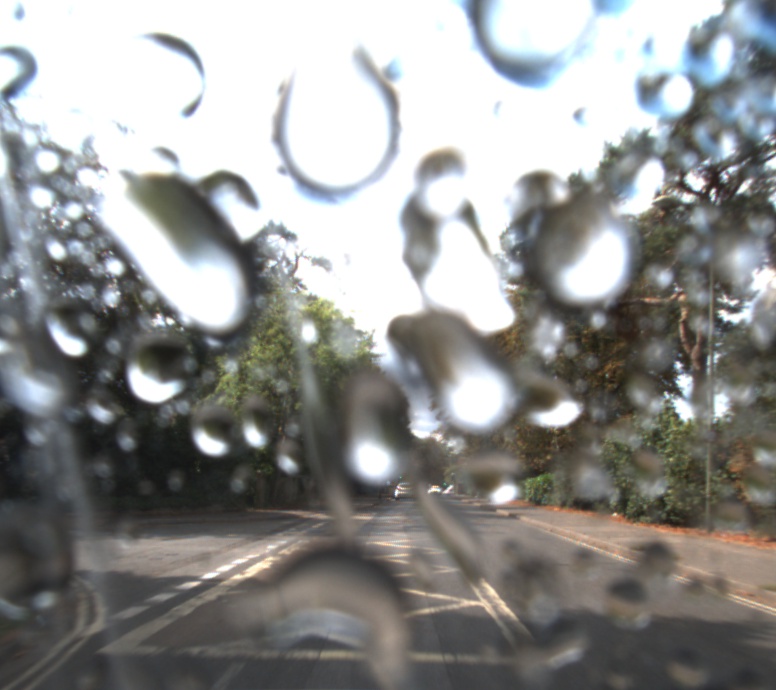}}%
\hfill %
{\includegraphics[width=0.248\textwidth]{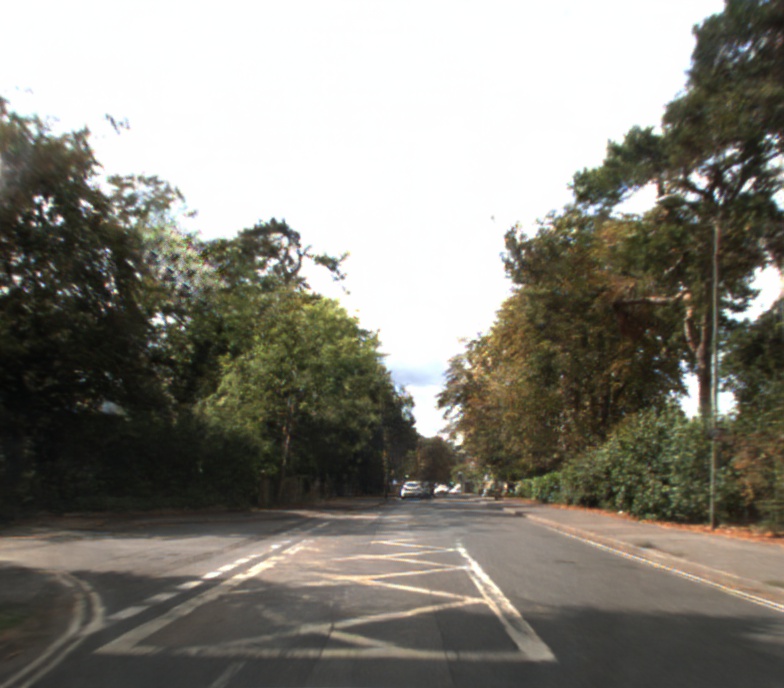}}%
\hfill
{\includegraphics[width=0.248\textwidth]{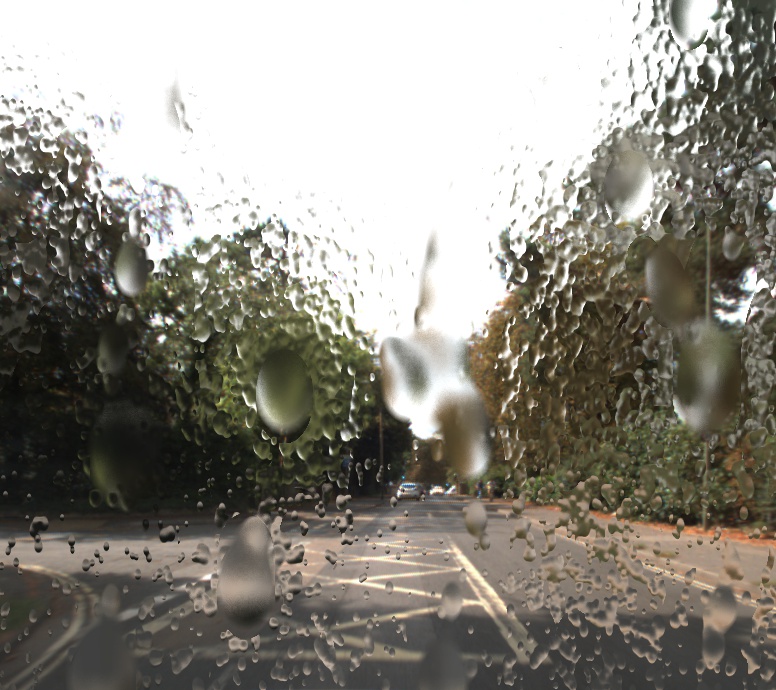}}%
\hfill %
{\includegraphics[width=0.248\textwidth]{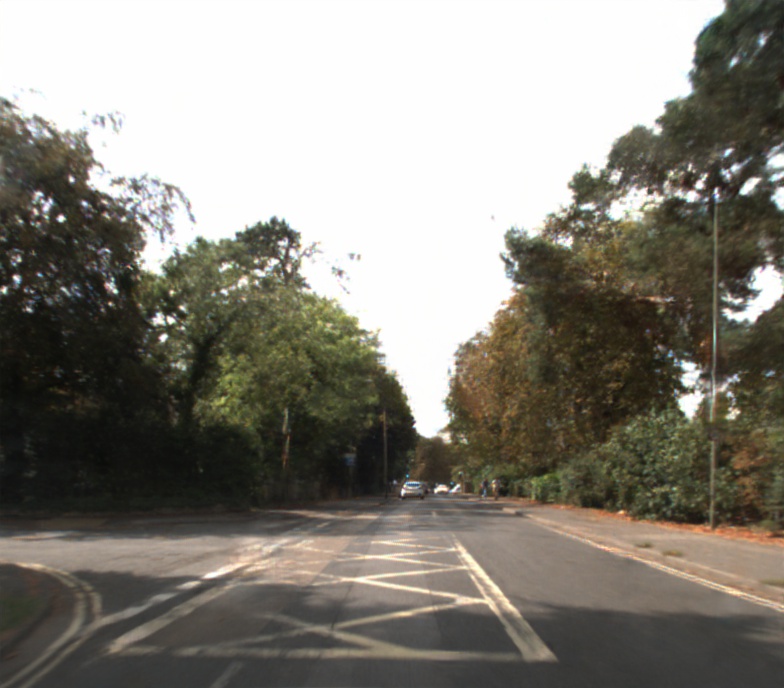}}%
\\ %
\vspace{0.5mm}

{\includegraphics[width=0.248\textwidth]{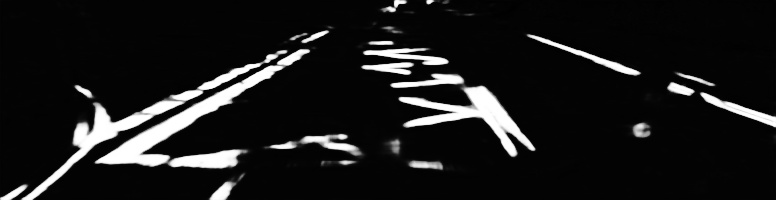}}%
\hfill %
{\includegraphics[width=0.248\textwidth]{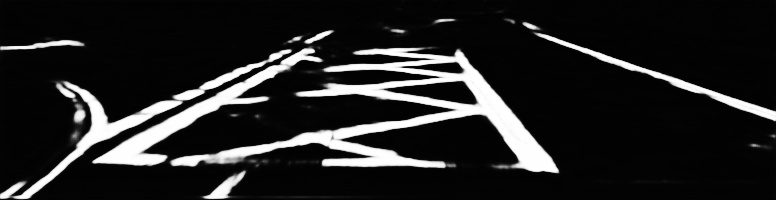}}%
\hfill %
{\includegraphics[width=0.248\textwidth]{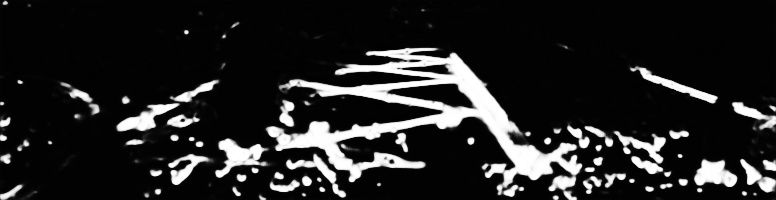}}%
\hfill %
{\includegraphics[width=0.248\textwidth]{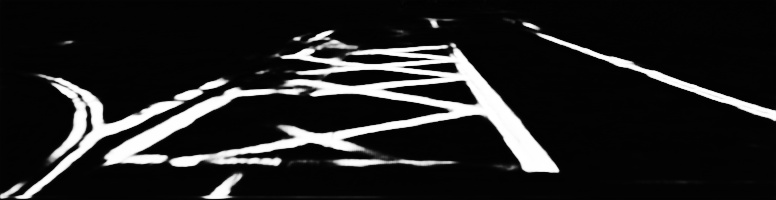}}%
\vspace{-3mm}
\caption{RobotCar road marking segmentation results.First column shows a RobotCar(R) real rain image and segmentation result. Second column shows the derained real rain image and segmentation result. Third column shows a RobotCar(S) computer-generated rain image and segmentation result. Fourth column shows the derained computer-generated rain image and segmentation result. }
\label{fig:rm}
\vspace{-3mm}
\end{figure*}

\begin{figure*}[!htbp]
\centering

{\includegraphics[width=0.498\textwidth]{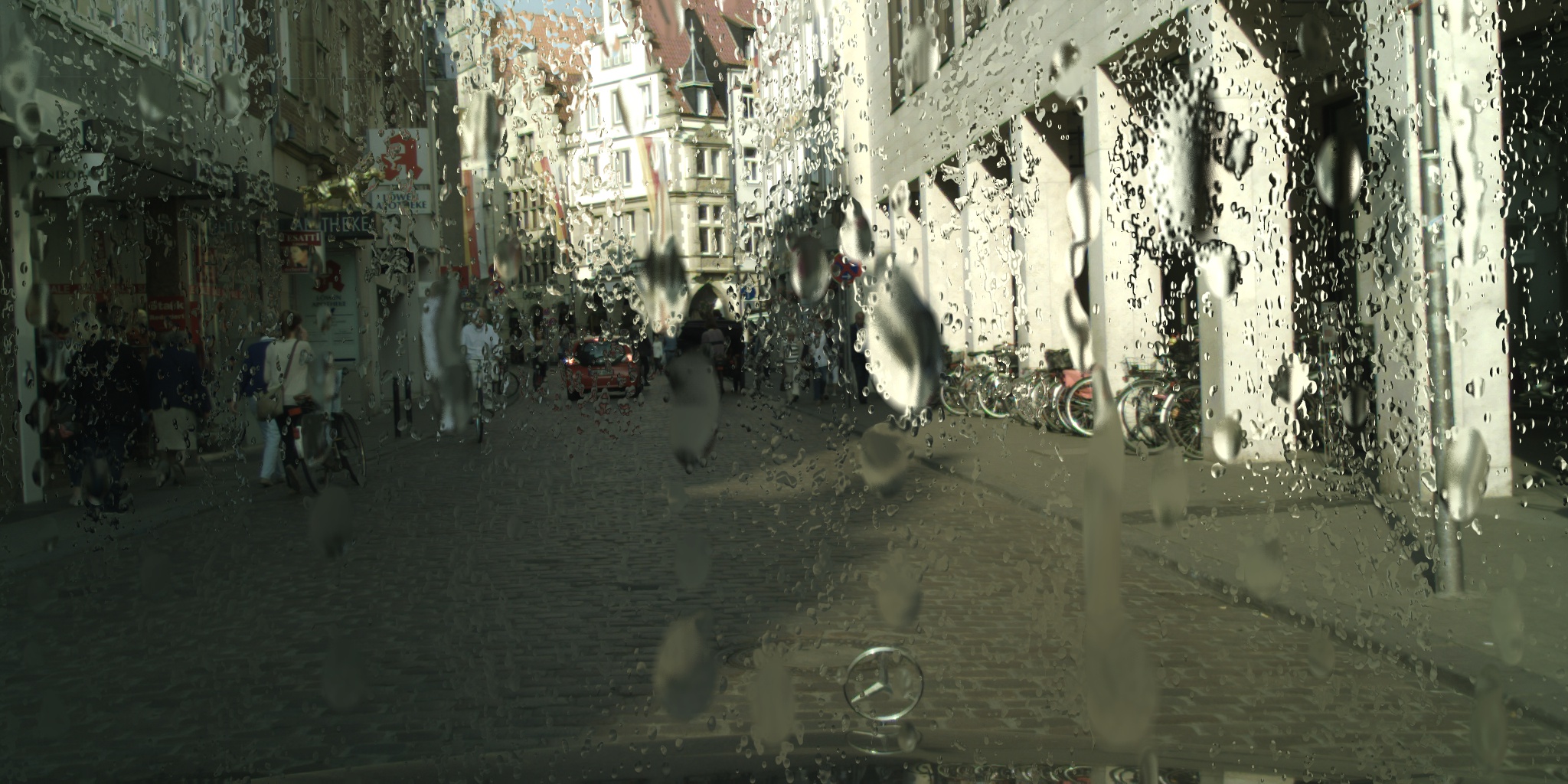}}%
\hfill %
{\includegraphics[width=0.498\textwidth]{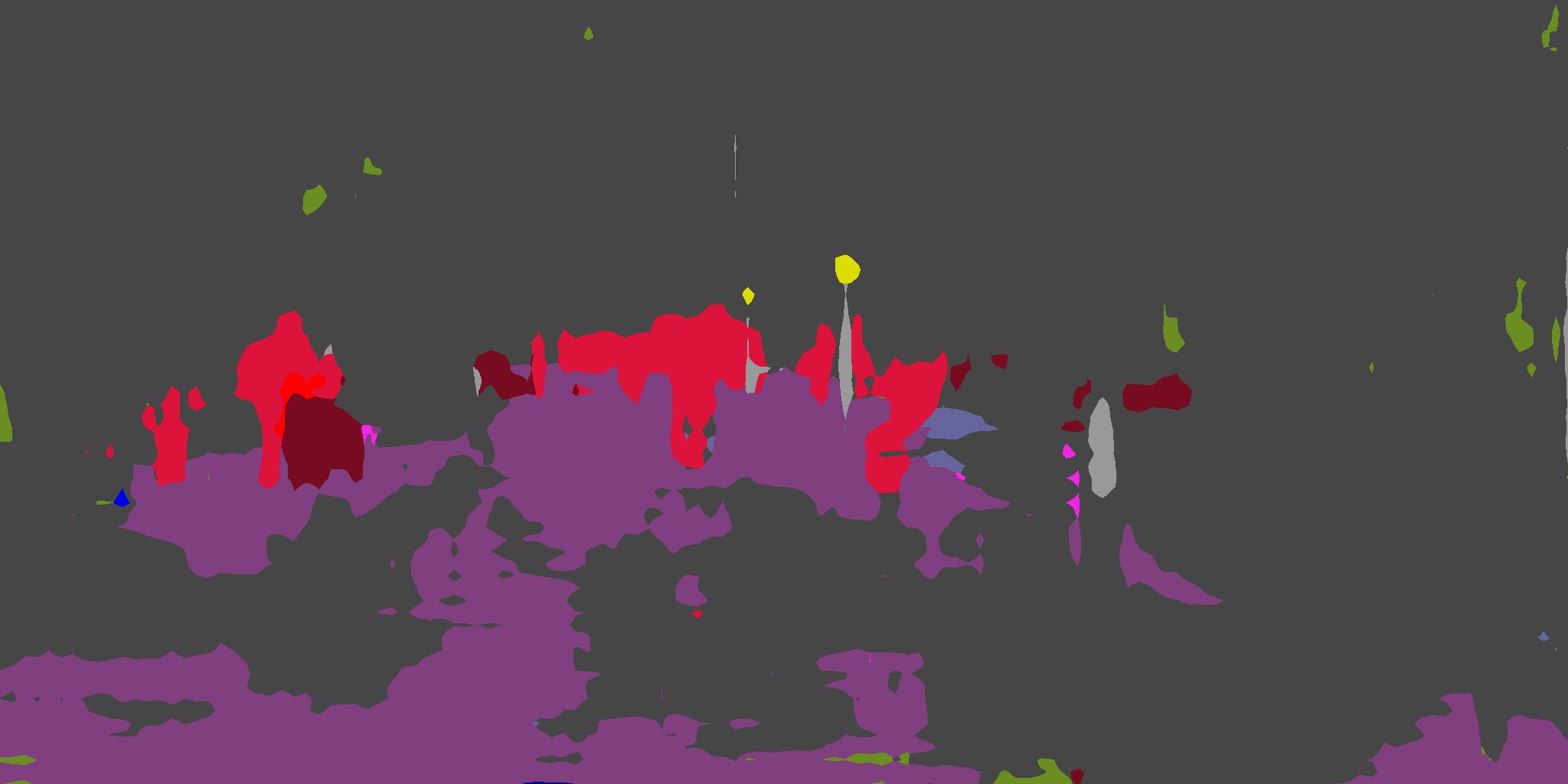}}%
\\ %
\vspace{0.5mm}

{\includegraphics[width=0.498\textwidth]{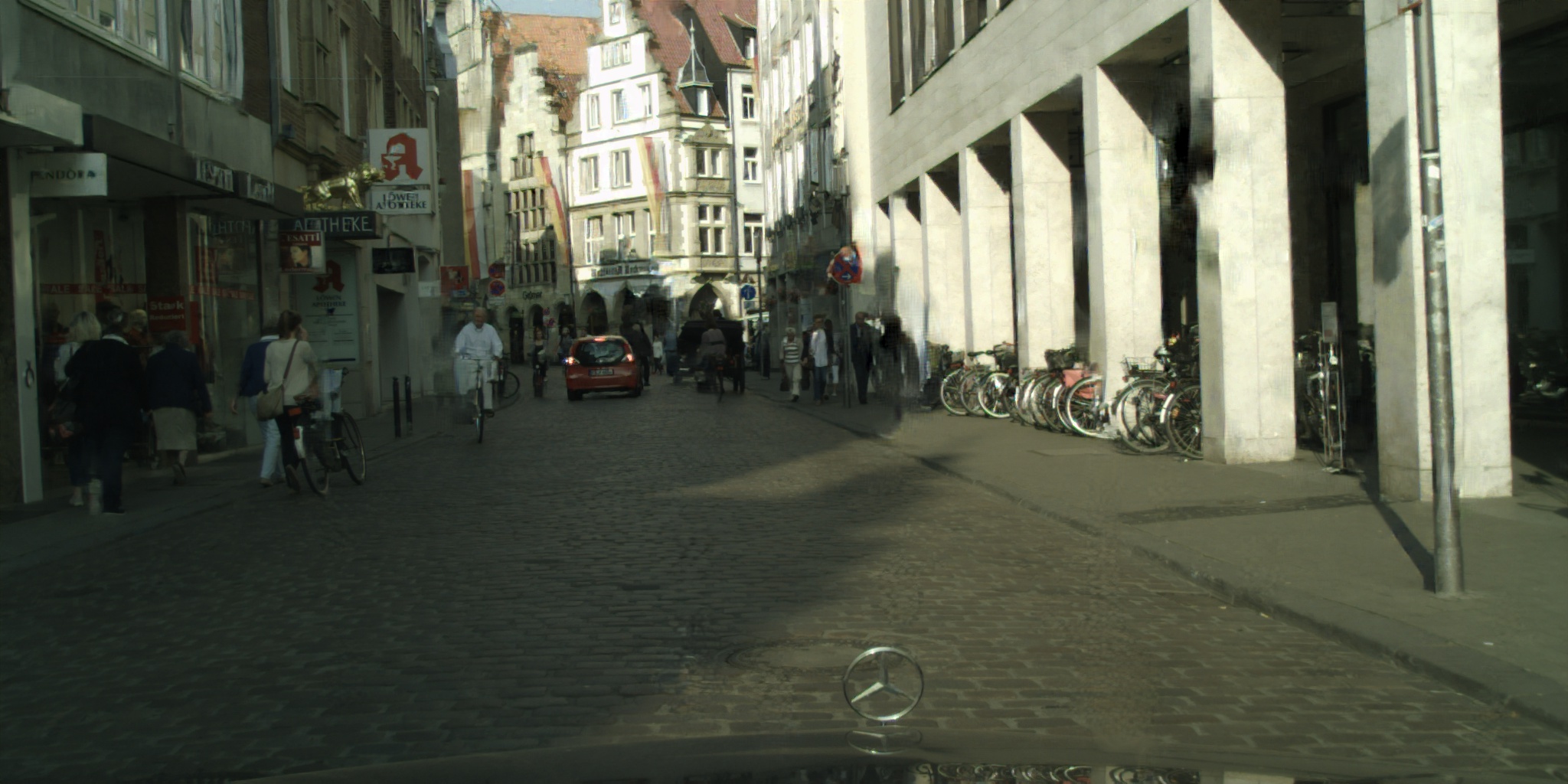}}%
\hfill %
{\includegraphics[width=0.498\textwidth]{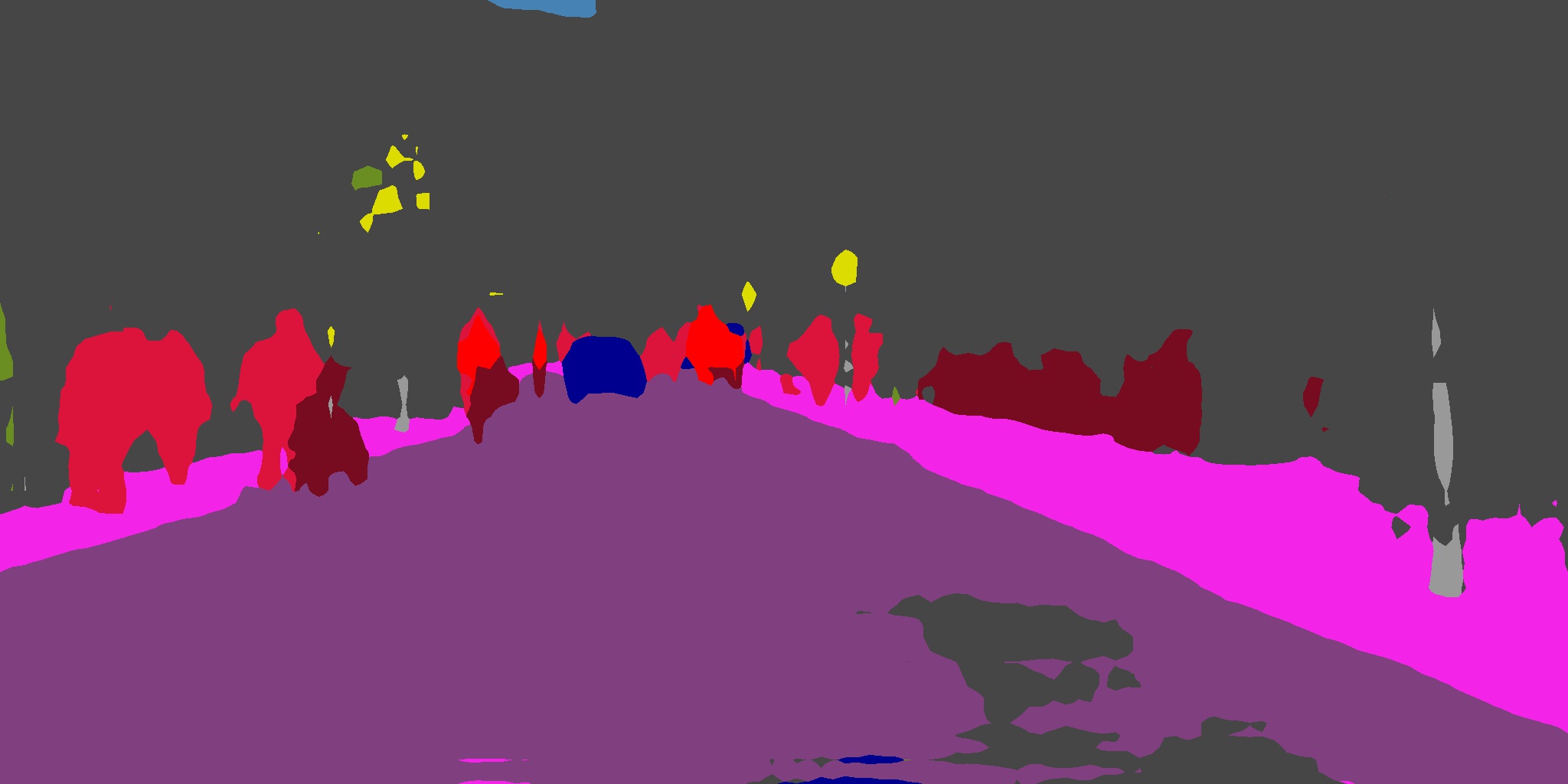}}%
\vspace{-2mm}
\caption{Cityscapes semantic segmentation results. The first row shows a rainy image on the left and its corresponding semantic segmentation on the right. The second row shows the derained image on the left and its corresponding semantic segmentation on the right.}
\label{fig:city}
\vspace{-3mm}
\end{figure*}

\begin{figure*}[!htbp]
\centering
 \includegraphics[width=\textwidth]{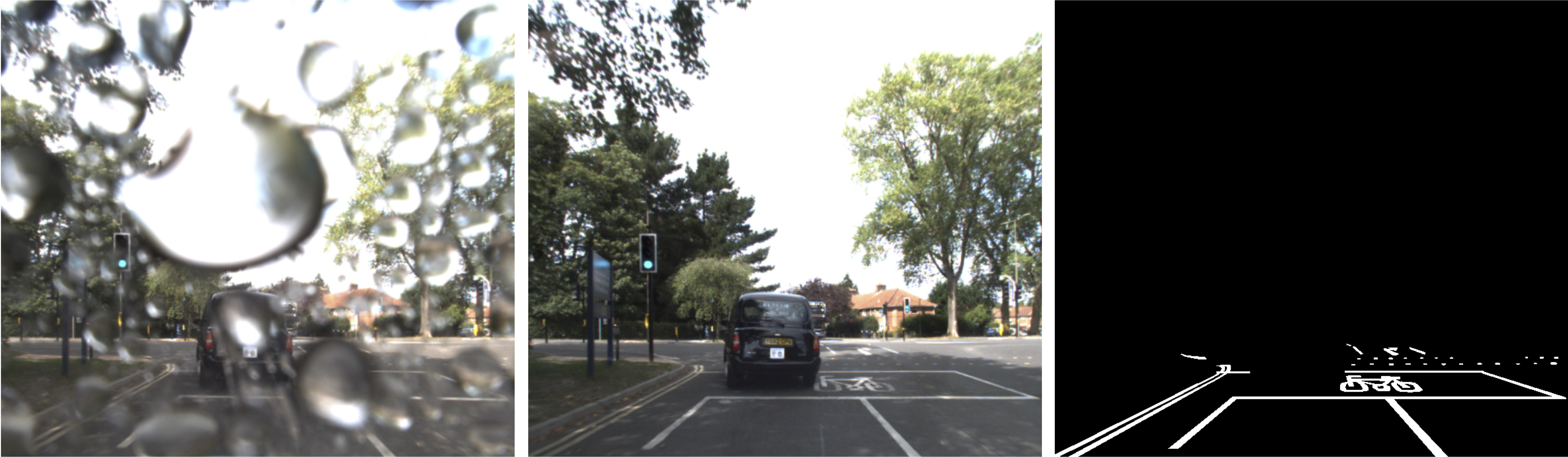}
\caption{An example from our stereo dataset. The image on the left is produced by the left lens, which is affected by water drops. The image in the middle is produced by the dry right hand lens. The image on the right is the road marking segmentation ground truth. }
\label{fig:dataset_example}
\vspace{-5mm}
\end{figure*}

\subsection{Losses}

Similar to \cite{CycleGAN2017}, we apply an adversarial loss through a discriminator on the output of the generator. This loss is formulated as:
\begin{equation}
 \mathcal{L}_{adv}= (D(G(I_{\mathrm{rainy}}))-1)^2.
\end{equation}

The discriminator is trained to minimize the following loss:
\begin{equation}
 \mathcal{L}_{disc}= (D(I_{\mathrm{clear}})-1)^2 + (D(I_{\mathrm{de-rained}}))^2,
\end{equation} where $I_{\mathrm{derained}}$ is sampled from a pool of previously derained images.

The perceptual loss \cite{johnson2016perceptual} is applied between the label and reconstructed image:
\begin{equation}
 \mathcal{L}_{\mathrm{perc}}=\sum_{i=1}^{n_{VGG}}\frac{1}{w_{i}^{perc}}{\lVert VGG(I_{\mathrm{clear}})_{i} - VGG(G(I_{\mathrm{rainy}}))_{i} \rVert}_{1},
\end{equation} where $n_{VGG}$ represents the number of VGG layers that are used to compute the loss and $w_{i}^{perc}=2^{(n_{VGG}-i)}$ weighs the importance of each layer.

Additionally, a multi-scale discriminator feature loss  \cite{wang2018highresolution} is applied between the label and reconstructed image:
\begin{equation}
 \mathcal{L}_{\mathrm{msadv}}=\sum_{i=1}^{n_{ADV}}\frac{1}{w_{i}^{adv}}{\lVert D(I_{\mathrm{clear}})_{i} - D(G(I_{\mathrm{rainy}}))_{i} \rVert}_{1},
\end{equation} where $n_{ADV}$ represents the number of discriminator layers that are used to compute the loss and $w_{i}^{adv}=2^{(n_{ADV}-i)}$ weighs the importance of each layer.

The complete generator objective $\mathcal{L}_{\mathrm{gen}}$ becomes:
\begin{multline}
 \mathcal{L}_{\mathrm{gen}}=\lambda_{\mathrm{adv}}*\mathcal{L}_{\mathrm{adv}} + \lambda_{\mathrm{perc}}*\mathcal{L}_{\mathrm{perc}} + \lambda_{\mathrm{msadv}} * \mathcal{L}_{\mathrm{msadv}}.
\end{multline}

Each $\lambda$ term is a hyperparameter that weights the importance of each term of the loss equation.

We wish to estimate the generator function $G$ such that:
\begin{equation}
 G = \underset{G,D}{\arg\min} \mathcal{L}_{\mathrm{gen}} + \mathcal{L}_{\mathrm{disc}}.
\end{equation}

In the following section we describe how the network is trained to minimise the above losses.

\section{Experimental Setup}\label{sec:experimental-setup}
\subsection{Stereo rain dataset}\label{sec:rainmaker}
\begin{figure}[!htbp]
\centering
 \includegraphics[trim={0cm 0.5cm 0cm 0cm},width=\columnwidth]{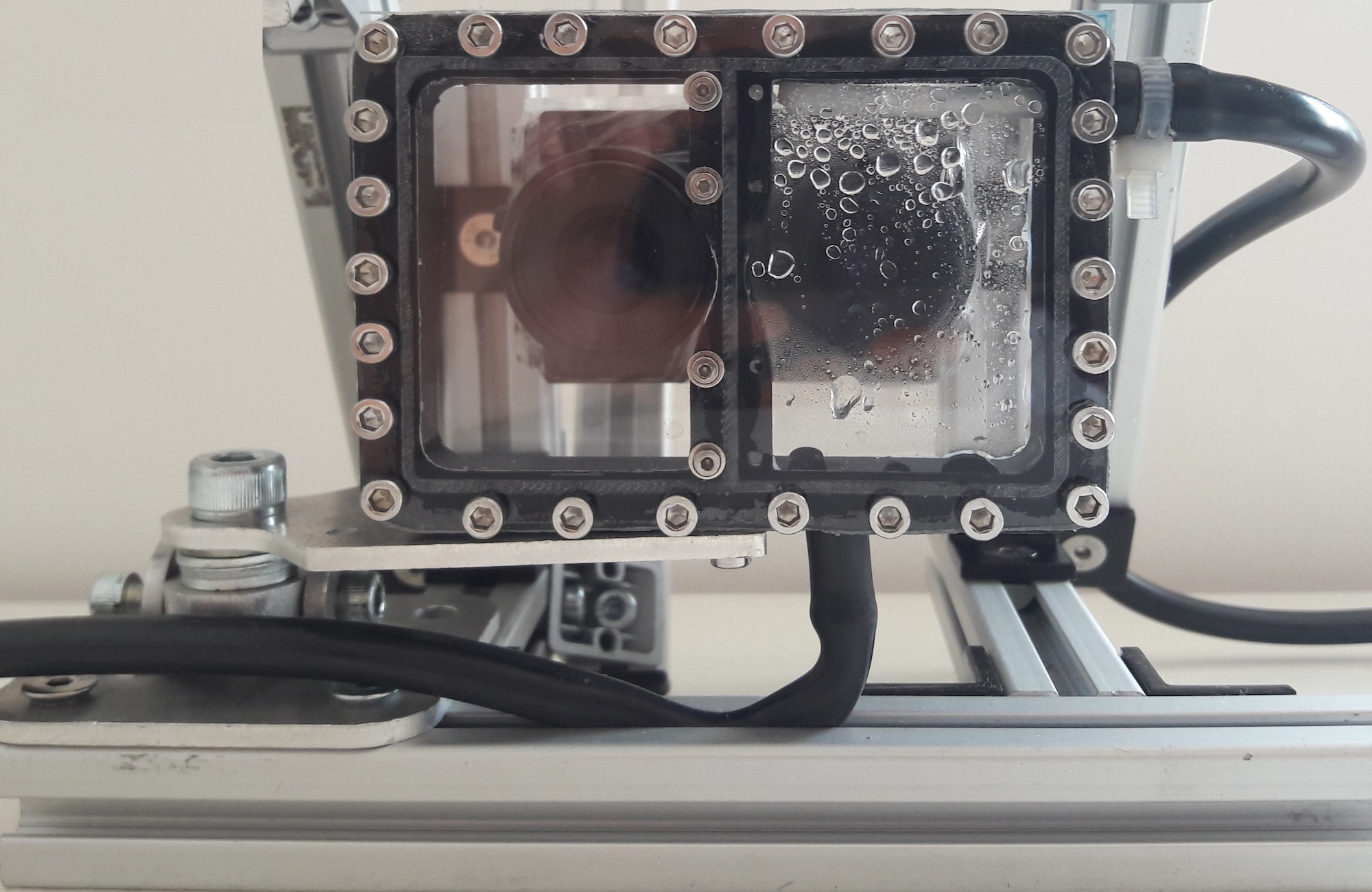}
\caption{Our small-baseline stereo camera setup. A bi-partite chamber with acrylic clear panels is placed in front of the lenses, with the left-hand section being kept dry at all times, while the right-hand section is sprayed with water droplets using an internal nozzle. }  
\label{fig:setup}
\end{figure}

In this section we present the hardware used to record our narrow-baseline stereo dataset that allows one lens to be affected by real water droplets while keeping the other lens clear. The camera setup is shown in Figure \ref{fig:setup}. A 3D-printed bi-partite chamber is sandwiched between two acrylic clear panels and placed in front of the two lenses, with the left-hand section of the chamber being kept dry at all times, while the right-hand section is sprayed with water droplets using an internal nozzle fitted at the top of the chamber. The angle of this chamber with respect to the axes of the cameras can be modified to simulate a slanted windscreen or enclosure, and the distance from the lenses can be increased or decreased accordingly to replicate different levels of focus or blur on the droplets. 

The nozzle spans the entire width of the right chamber and is capable of producing water droplets with a diameter between 1mm and 8mm, as well as streaks of water. This variability is achieved by modulating the water pressure using a number of pulse width modulation regimes. The water is drained from the bottom of the chamber and is returned to a storage tank for recirculation.
The cameras used are Point Grey Grasshopper 2 with 4.5\,mm F/1.4 lenses, a baseline of 29\,mm and automatic synchronisation. The system is fully portable and the water is completely contained within the circuit formed by the right chamber, pump and tank.

We have collected approximately 50000 pairs of images by driving in and around the city of Oxford. The image pairs are undistorted, cropped and aligned. We have selected 4818 image pairs to form a training, validation and testing dataset. From the testing partition, we have created ground truth road marking segmentations for 500 images. An example from our dataset is shown in Figure \ref{fig:dataset_example}.

Compared to the painstakingly-collected dataset of \cite{Qian2018}, our setup is a set-and-forget approach: once the stereo camera has been mounted on a vehicle, it is trivial to collect large amounts of well-synchronised and well-aligned pairs of images.

\vspace{-2mm}

\subsection{Training}

We used a network training regimen similar to \cite{wang2018highresolution}. 
For each iteration we first trained the discriminator on a clear image and a de-rained image from a previous iteration with the goal of minimizing $\mathcal{L}_{\mathrm{disc}}$, and then trained the generator on rainy input images to minimize $\mathcal{L}_{\mathrm{gen}}$. 
We used the Adam solver \cite{kingma2014adam} with an initial learning rate set at 0.0002, a batch size of 1, $\lambda_{\mathrm{adv}}=1$, $\lambda_{\mathrm{perc}}=1$ and $\lambda_{\mathrm{msadv}}=1$.

\begin{table*}[t]
    \caption{Road marking segmentation results}
    \vspace{-5mm}
	\begin{center}
		\begin{tabular}{@{} |c| C{0.4cm}*{3}{C{0.6cm}} C{0.4cm}*{3}{C{0.6cm}} C{0.4cm}*{3}{C{0.6cm}} C{0.4cm}*{3}{C{0.6cm}} @{}}
			\hline
			\multirow{3}*{\bf Dataset} & \multicolumn{4}{c|}{\bf REFERENCE(CLEAR)}& \multicolumn{4}{c|}{\bf RAINY} & \multicolumn{4}{c|}{\bf AUGM.} & \multicolumn{4}{c|}{\bf DERAINED}\\
            & Prec.&Rec.&F1&\multicolumn{1}{C{0.7cm}|}{IOU}& Prec.&Rec.&F1&\multicolumn{1}{C{0.7cm}|}{IOU}& Prec.&Rec.&F1&\multicolumn{1}{C{0.7cm}|}{IOU}& Prec.&Rec.&F1&\multicolumn{1}{C{0.7cm}|}{IOU} \\ \hline
            RobotCar(R) & 0.627&0.918&0.734&\multicolumn{1}{c|}{0.594}& 0.512&0.628&0.550&\multicolumn{1}{c|}{0.396}& 0.486&0.807&0.593&\multicolumn{1}{c|}{0.434}& 0.603&0.841&0.689&\multicolumn{1}{c|}{0.544} \\ \hline
            RobotCar(S) & 0.627&0.918&0.734&\multicolumn{1}{c|}{0.594}& 0.364&0.595&0.437&\multicolumn{1}{c|}{0.287}& 0.654&0.770&0.690&\multicolumn{1}{c|}{0.541}& 0.661&0.816&0.715&\multicolumn{1}{c|}{0.569} \\ \hline
            CamVid(S) & 0.5763&0.9269&0.6992&\multicolumn{1}{c|}{0.5512}& 0.3533&0.5762&0.4248&\multicolumn{1}{c|}{0.2787}& 0.4568&0.7707&0.5635&\multicolumn{1}{c|}{0.4051}& 0.5198&0.7553&0.6030&\multicolumn{1}{c|}{0.4438} \\ \hline
		\end{tabular}
	\end{center}
	\label{tab:SegmRes}
\end{table*}

\begin{table}[t]
	\begin{center}
	\caption{Cityscapes Semantic segmentation results}
		\begin{tabular}{@{} |c| C{0.8cm} @{}}
			\hline
			\multirow{1}*{\bf Cityscapes Model vs. Dataset}
			& \multicolumn{1}{C{0.8cm}|}{mIOU} \\ \hline
			CLEAR on CLEAR & \multicolumn{1}{c|}{0.692}\\ \hline
			RAINY on CLEAR & \multicolumn{1}{c|}{0.405} \\ \hline
			RAINY on AUGMENTED & \multicolumn{1}{c|}{0.611} \\ \hline
			DERAINED on CLEAR & \multicolumn{1}{c|}{0.651} \\ \hline
		\end{tabular}
	\end{center}
	\label{tab:semseg}
\end{table}

\subsection{Segmentation Tasks}

We used the trained generator $G$ to de-rain all of the rainy input images. To benchmark both the images with computer-generated water drops and the images with real water drops, in the context of road marking segmentation, we used the approach of \cite{bruls2018mark} which trains a U-Net to segment road markings in a binary way. To benchmark the computer-generated water drop images in the context of semantic segmentation, we used DeepLab v3 \cite{deeplabv3plus2018} which has achieved state-of-the-art performance on the Cityscapes dataset.

The generator runs at approximately $1$\,Hz for images with a resolution of $1280\times960$, and at approximately $3$ Hz for images with a resolution of $640\times480$ on an Nvidia Titan X GPU.

\section{Results}\label{sec:results}

We benchmark our results taking into consideration several metrics across several tasks, and also present results on the quality of the image reconstruction.

\subsection{Quantitative results}

Table \ref{tab:SegmRes} presents results for road marking segmentation, in the case of RobotCar with real water drops (R), RobotCar with computer-generated water drops (S) and CamVid with computer-generated water drops (S).
Our baseline is represented by the performance of clear images tested on models that were trained using clear images (REFERENCE). For both RobotCar (R), Robotcar (S), and the CamVid (S) datasets, the results show a severely degraded performance when testing rainy images on models that were trained using clear images (RAINY). Retraining the road marking segmentation models with a dataset augmented with rainy images will lead to an improvement in performance (AUGM). However, de-raining the images using our method and testing them on a model trained using clear images (DERAINED) restores the performance of the segmentation to levels that are close to the baseline recorded on clear images. Figure \ref{fig:camvid} shows road marking segmentation results on CamVid, before and after deraining. Figure \ref{fig:rm} shows road marking segmentation results on RobotCar(R)\&(S), before and after deraining.

As expected, re-training the segmentation model with a dataset that is augmented with rainy images helps to improve performance, however using a specialised de-raining preprocessing step significantly outperforms this approach, even when tested on a model trained exclusively with clear images. This is the expected advantage of having a model dedicated, in its entirety, to a specific image-to-image mapping task (de-raining), which narrows the variety of images fed to the segmentation task.

Table \ref{tab:semseg} presents results for semantic segmentation on the Cityscapes dataset. We benchmark 4 different combinations of models and datasets:
\begin{itemize}
    \item Cityscapes-clear images tested on a model trained using Cityscapes-clear images;
    \item Cityscapes-rainy images tested on a model trained using Cityscapes-clear images;
    \item Cityscapes-rainy images tested on a model trained using Cityscapes-clear and Cityscapes-rainy  images; and
    \item Cityscapes-derained(Cityscapes-rainy preprocessed using our deraining model) images tested on a model trained using Cityscapes-clear images.
\end{itemize}
Similar to the case of road marking segmentation, we notice the same severe degradation of performance when testing with rainy images (RAINY on CLEAR) as compared to the baseline (CLEAR on CLEAR). Again, the performance of derained images tested on a model trained using clear images (DERAINED on CLEAR) is significantly better than the performance of rainy images tested on a model trained using a dataset augmented with rainy images(RAINY on AUGMENTED). Figure \ref{fig:city} shows semantic segmentation results on Cityscapes, before and after deraining.

\subsection{Reconstruction results}

\begin{table}[t]
    \caption{Reconstruction results}
    \vspace{-5mm}
	\begin{center}
		\begin{tabular}{@{} |c| C{0.6cm}*{1}{C{0.6cm}} C{0.6cm} *{1}{C{0.6cm}} @{}}
			\hline
			\multirow{3}*{\bf Dataset} & \multicolumn{2}{c|}{\bf RAW}& \multicolumn{2}{c|}{\bf DERAINED} \\
			& PSNR&\multicolumn{1}{C{0.7cm}|}{SSIM}& PSNR &\multicolumn{1}{C{0.7cm}|}{SSIM} \\ \hline
			RobotCar-Rainy(R) & 13.02&\multicolumn{1}{c|}{0.5574}& 22.82&\multicolumn{1}{c|}{0.8188} \\ \hline
			RobotCar-Rainy(S) & 16.80&\multicolumn{1}{c|}{0.6134}& 25.17&\multicolumn{1}{c|}{0.8699} \\ \hline
			CamVid-Rainy(S) & 16.89&\multicolumn{1}{c|}{0.6064}& 22.11&\multicolumn{1}{c|}{0.7524} \\ \hline
			Qian et al.\cite{Qian2018}(R) & 24.09&\multicolumn{1}{c|}{0.8518}& 31.55&\multicolumn{1}{c|}{0.9020} \\ \hline
		\end{tabular}
	\end{center}
	\label{tab:recresults}
\end{table}

\begin{table}[t]
    \caption{Reconstruction quality comparison to state of the art}
    \vspace{-5mm}
	\begin{center}
		\begin{tabular}{@{} |c| C{0.8cm} C{0.6cm} @{}}
			\hline
			\multirow{2}*{\bf Model vs. Dataset} & \multicolumn{2}{c|}{\bf Dataset from \cite{Qian2018}} \\
			& \multicolumn{1}{C{0.8cm}}{PSNR}&\multicolumn{1}{C{0.6cm}|}{SSIM} \\ \hline
			Original & \multicolumn{1}{c|}{24.09}&\multicolumn{1}{c|}{0.8518} \\ \hline
			Eigen13\cite{EigenDerain} & \multicolumn{1}{c|}{28.59}&\multicolumn{1}{c|}{0.6726} \\ \hline
			Pix2Pix\cite{pix2pixold} & \multicolumn{1}{c|}{30.14}&\multicolumn{1}{c|}{0.8299} \\ \hline
			Qian et al.(no att.)\cite{Qian2018} & \multicolumn{1}{c|}{30.88}&\multicolumn{1}{c|}{0.8670} \\ \hline
			Qian et al.(full att.)\cite{Qian2018} & \multicolumn{1}{c|}{31.51}&\multicolumn{1}{c|}{\textbf{0.9213}} \\ \hline
			Ours(no att.) & \multicolumn{1}{c|}{\textbf{31.55}}&\multicolumn{1}{c|}{0.9020} \\ \hline
		\end{tabular}
	\end{center}
	\label{tab:Qualitative}
\end{table}

Table \ref{tab:recresults} presents results on the quality of the image reconstruction using two widely used image-quality metrics, PSNR and SSIM. We benchmark our model on our real-world RobotCar-Rainy (R) dataset, RobotCar-Rainy with computer-generated rain (S), CamVid-Rainy with computer-generated rain (S), and on the dataset provided by \cite{Qian2018}. The RAW column shows the quality of the rainy images, while the DERAINED column shows the quality of the de-rained images, all relative to their clear ground truth. We show that in all cases, de-raining the rain-affected images using our preprocessor significantly increases the quality of the images, as compared to the reference case where raw rainy images are used. Both the real-world  rainy dataset images and the images with computer-generated rain are significantly more degraded than the rainy images provided by \cite{Qian2018}, as seen in column RAW.

Table \ref{tab:Qualitative} presents reconstruction results on the reference rainy dataset provided by \cite{Qian2018}. We show that we achieve state-of-the-art PSNR reconstruction results on images affected by real water drops and only slightly lower SSIM, while, in contrast to \cite{Qian2018}, not requiring an attention \cite{attention} mechanism, which simplifies and speeds up inference and training.

\section{Conclusions}\label{sec:conclusions}

We have presented a system that restores performance of images affected by adherent raindrops on important segmentation tasks. Our results show that road marking segmentation, an important task for autonomous driving systems, is severely affected by adherent rain and that performance can be restored by first running the images through a de-raining preprocessor. Similarly, we show the same reduction and restoration of performance in the case of semantic segmentation, a task that is important in many fields. Additionally, we produce state-of-the-art results in terms of the quality of image restoration, while being able to run in real time. Finally, our system processes the image streams outside of the segmentation pipeline, either offline or online, and hence can be used naturally as a front end to many existing systems. 
The dataset will be made available at \url{https://ciumonk.github.io/RobotCar-rainy/}, along with a video describing our results at  \url{https://ciumonk.github.io/RobotCar-rainy/video.html}.

\section{Future work}\label{sec:future}

Future work may involve designing a mechanism for producing computer-generated rain that is indistinguishable from real rain in terms of its usefulness in training models that quantitatively rather than qualitatively improve performance on image-based tasks.
\section{Acknowledgements}\label{sec:acknowledgements}

This work was supported by Oxford-Google DeepMind Graduate Scholarships and Programme Grant EP/M019918/1. The authors wish to thank Valentina Musat for labelling the road markings in our dataset.

\bibliographystyle{IEEEtran}
{\footnotesize
\bibliography{icrasynthetics2017}}

\end{document}